\documentclass[journal]{IEEEtran}
\usepackage{ifpdf}
\usepackage{threeparttable}
\usepackage{amsfonts}
\usepackage{makecell}
\usepackage{nicefrac}
\usepackage{xcolor}
\usepackage{tabularx}
\usepackage{cite}
\usepackage[backref]{hyperref}
\usepackage{xspace}
\newcommand{\ourapproach}{\textsc{HRNN}\xspace}  % model name if you have.
\usepackage{threeparttable}
\ifCLASSINFOpdf
  \usepackage[pdftex]{graphicx}
  \graphicspath{{../pdf/}{../jpeg/}}
  \DeclareGraphicsExtensions{.pdf,.jpeg,.png}
\else
  \usepackage[dvips]{graphicx}
  \graphicspath{{../eps/}}
  \DeclareGraphicsExtensions{.eps}
\fi
\usepackage{amsmath}
\usepackage{algorithmic}
\usepackage{multirow}
\usepackage{array}
\usepackage{footnote}
\ifCLASSOPTIONcompsoc
  \usepackage[caption=false,font=normalsize,labelfont=sf,textfont=sf]{subfig}
\else
  \usepackage[caption=false,font=footnotesize]{subfig}
\fi
\usepackage{caption}
\usepackage{fixltx2e}
\usepackage{dblfloatfix}
\usepackage{url}
\usepackage{authblk}
\begin{document}
\title{Identifying Electrocardiogram Abnormalities Using a Handcrafted-Rule-Enhanced Neural Network}

\author{Yuexin Bian, Jintai Chen, Xiaojun Chen, \\Xiaoxian Yang$^{*}$, Danny Z. Chen,~\IEEEmembership{Fellow,~IEEE}, Jian Wu$^*$,~\IEEEmembership{Member,~IEEE}
\thanks{Yuexin Bian and Jintai Chen are with the College of Computer Science and Technology, Zhejiang University, Hangzhou, 310027, China (e-mails: alwaysbyx@gmail.com, jtchen721@gmail.com).}
\thanks{Xiaojun Chen is with the RealDoctor AI Research Centre, Zhejiang University, Hangzhou, 310000, China. (e-mail: chenxiaojun@zju.edu.cn)}
\thanks{Xiaoxian Yang is with the School of Computer and Information Engineering, Shanghai Polytechnic University, China (e-mail: xxyang@sspu.edu.cn).}\thanks{Danny Z. Chen is with the Department of Computer Science and Engineering, University of Notre Dame, Notre Dame, IN 46556, USA (e-mail: dchen@nd.edu).}\thanks{Jian Wu is with the Second Affiliated Hospital School of Medicine and School of Public Health, Zhejiang University, Hangzhou 310058, China (e-mail: wujian2000@zju.edu.cn).}\thanks{$^*$Corresponding authors.}}

% The paper heanders
% \markboth{Journal of \LaTeX\ Class Files,~Vol.~14, No.~8, August~2015}%
% {Shell \MakeLowercase{\textit{et al.}}: Bare Demo of IEEEtran.cls for IEEE Journals}



\maketitle

\begin{abstract}
A large number of people suffer from life-threatening cardiac abnormalities, and electrocardiogram (ECG) analysis is beneficial to determining whether an individual is at risk of such abnormalities. Automatic ECG classification methods, especially the deep learning based ones, have been proposed to detect cardiac abnormalities using ECG records, showing good potential to improve clinical diagnosis and help early prevention of cardiovascular diseases. However, the predictions of the known neural networks still do not satisfactorily meet the needs of clinicians, and this phenomenon suggests that some information used in clinical diagnosis may not be well captured and utilized by these methods. In this paper, we introduce some rules into convolutional neural networks, which help present clinical knowledge to deep learning based ECG analysis, in order to improve automated ECG diagnosis performance. Specifically, we propose a Handcrafted-Rule-enhanced Neural Network (called \ourapproach) for ECG classification with standard 12-lead ECG input, which consists of a rule inference module and a deep learning module. Experiments on two large-scale public ECG datasets show that our new approach considerably outperforms existing state-of-the-art methods. Further, our proposed approach not only can improve the diagnosis performance, but also can assist in detecting mislabelled ECG samples. Our codes are available at \href{https://github.com/alwaysbyx/ecg_processing}{https://github.com/alwaysbyx/ecg\_processing}.
% making up the shortcomings of complete deep learning methods to detect mislabelled samples.
\end{abstract} 


\IEEEpeerreviewmaketitle

\section{Introduction}~\label{intro}
\IEEEPARstart{E}{lectrocardiogram} (ECG) is a type of commonly-used test in clinical practice for diagnosing patients suffered from cardiac abnormalities. Over 300 million ECG records are produced worldwide each year~\cite{holst1999confident}, which are a heavy burden for manual ECG diagnosis. For example, in China, around 250 million individuals take ECG tests each year, but only around 36,000 proficient doctors are engaged in analyzing such ECG data~\cite{2017panfeng}. Thus, there is a clear gap between the supply and demand in clinical ECG diagnosis.
\begin{figure}[t]
\centering
\includegraphics[width=1.\linewidth]{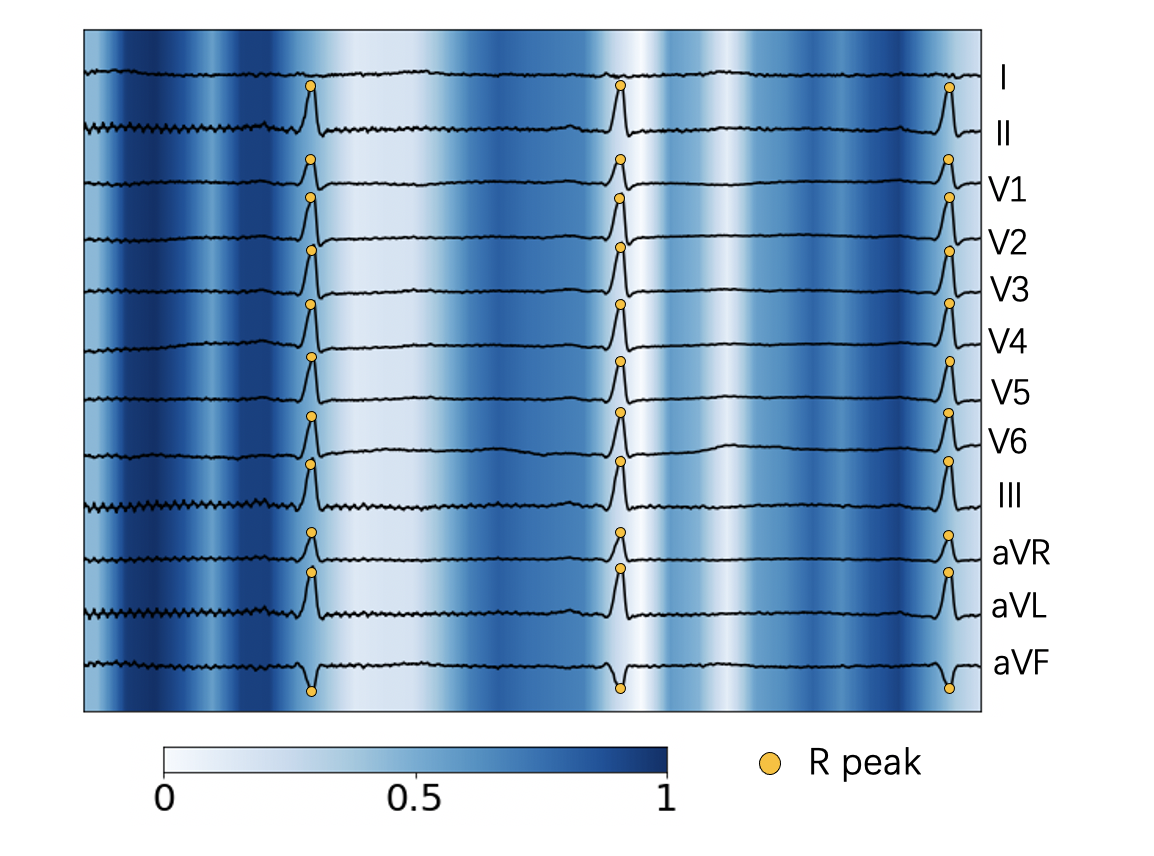}
\caption{Illustrating a saliency heat map for a case belonging to Counterclockwise Rotation (CCR, a type of cardiac abnormality). The ECG segments of interest are obtained by 1D ResNet-34 with Grad-CAM~\cite{gradcam}, and are visualized after being scaled with importance and normalized to 0--1. Yellow dots refer to the R peaks of ECG.}\label{fig:gradcam}
\end{figure}

Recently, artificial intelligence (AI) methods have been revolutionizing various tasks~\cite{he2016deep,wei2019adversarial,wei2020lifelong,dai2019transformer,wei2021incremental}, including diagnosis practices~\cite{kiranyaz2015real,pyakillya2017deep,saadatnejad2019lstm,hughes2021performance,chen2020flow,chen2021electrocardio} and provided effective assistance in automated signal analysis. Compared with doctors' (manual) analysis of ECG, such methods can make cardiac abnormality diagnosis more efficient.
%helping improve ECG diagnosis applications.
Among them, many machine learning methods were proposed for automated ECG diagnosis. In early years, various traditional methods employing decision trees~\cite{rodriguez2005real}, SVM~\cite{celin2018ecg}, random forests~\cite{ansari2017review}, and Bayesian networks~\cite{venkatesan2018novel} were applied to classify ECG signals, but did not yield satisfactory performances. Recently, deep learning approaches have drastically improved performances of various recognition tasks, including automatic ECG diagnosis~\cite{pyakillya2017deep,chen2020flow,bizopoulos2018deep,mathews2018novel, hannun2019cardiologist,luo2019multi}. Deep learning methods for ECG can be roughly divided into three types, graph based~\cite{wang2020weighted,jiang2020diagnostic,liu2020ecg}, recurrent neural network (RNN) based~\cite{yildirim2018novel, hou2019lstm, gao2019effective}, and convolution based~\cite{yao2020multi,chen2020flow,chen2021electrocardio,hannun2019cardiologist} methods. Specifically, graph based neural networks aim to capture the dependencies among cardiac abnormalities, since many ECG cases belong to multiple categories (abnormality types). RNN based methods treat ECG signals as time series and perform temporal feature extraction. Convolution based methods process ECG data as a special case of images, and often achieve better results~\cite{hannun2019cardiologist}.

\begin{figure*}[t]
\centering
\includegraphics[width=1.\linewidth]{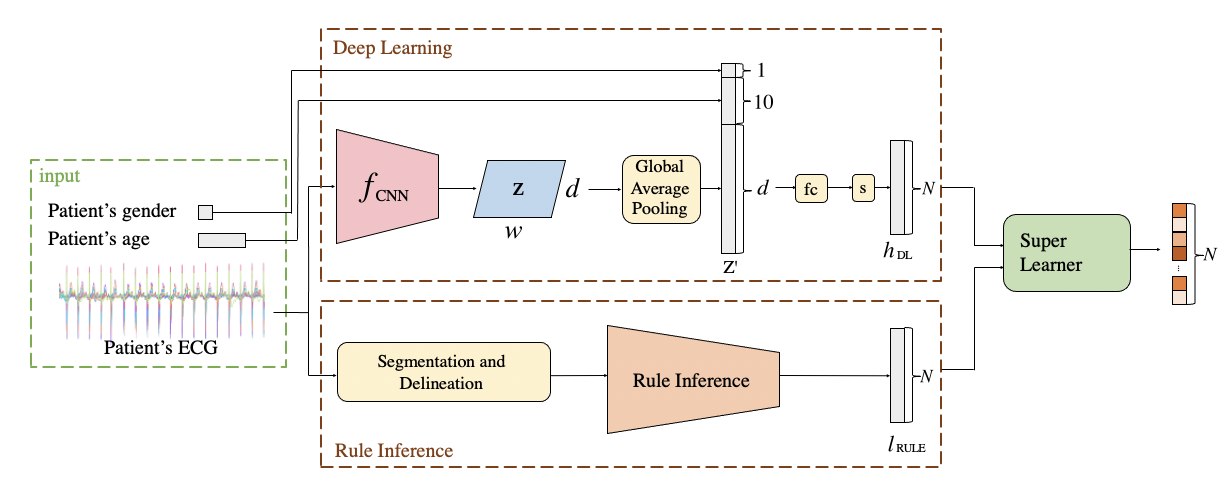}
\caption{The overall framework of our \ourapproach for multi-label ECG abnormality classification. $N$ denotes the types of abnormalities. We transform the patient's gender and age into a scalar and a vector of size 10, respectively, and concatenate them with the final $d$-dimensional feature vector extracted by the CNN with the global average pooling operation (``fc'' indicates a fully-connected layer and ``s'' indicates a sigmoid layer).}\label{frame}
\end{figure*}

%While many previous studies of ECG classification were based on neural network methods~\cite{bortolan1993diagnostic, gautam2016neural, li2017classification}, 
%some evidences~\cite{islam2020much} actually pointed out that common neural networks cannot exactly focus on the fine-grained and detailed information like locations
%\textcolor{red}{some commonly-used and inevitable 
%neural operations including normalization and pooling operation suggest that neural network methods may not be able to exactly focus on the fine-grained and detailed information like position information, which hinders the ECG examination. 
In clinical practice, clinicians often analyze ECG records in two main aspects: (1) experienced doctors may analyze whether the shapes of some key segments are normal, including P waves, T waves, and QRS complexes; (2) clinicians quantitatively analyze voltages and duration of certain waves. However, the existing neural network based methods do not seem to be able to capture the key segment voltages well.
%concisely. 
For example, we performed a classification study on ECG signals, and observed some inconsistency between the segments taken seriously by clinicians and those by neural networks. An example for this is given in Fig.~\ref{fig:gradcam}. A clinical diagnostic criterion for the Counterclockwise Rotation (CCR) abnormality is to identify the presence of an R peak shift in the lead V1-V6. In Fig.~\ref{fig:gradcam}, the blue parts mark the critical segments of a CCR case obtained by a trained 1D ResNet-34~\cite{he2016deep} with the Grad-CAM approach~\cite{gradcam}. One can see that the 1D ResNet-34 does not even focus on the R peaks (as marked in Fig.~\ref{fig:gradcam}), which is obviously not consistent with the clinical criterion for CCR. Similar inconsistency can be found on detecting other abnormalities on ECG. For a specific abnormality, clinicians can determine whether it presents by analyzing the corresponding key segment (e.g., an ST segment), but neural network based methods may give weights to every point in an ECG signal and do not even care about the key segment focused by clinicians.
% It is obvious that some important features for clinical knowledge are not captured by neural networks, 
% ( the voltage amplitudes and the locations of certain waves) are important for diagnosis from the perspective of clinical knowledge, and ECG analysis was proved to be a fine-grained task~\cite{Marriott0Marriott}. However, some common and inevitable neural operations for the comprehensive semantics abstraction, such as the normalization (e.g., the Batch Normalization~\cite{bn}), and pooling operation (e.g., the average pooling), are not good at preserving such fine-grained information. Previous work~\cite{islam2020much} revealed that the current convolutional neural networks can only learn a certain degree of location information in an implicit way, but such kind of location information is not sufficient to meet the more complex requirements in complex tasks. Also, the analyses in capsule neural networks~\cite{sabour2017dynamic,ribeiro2020introducing,hinton2018matrix,chen2021receptor} pointed out that the down-sampling operations in neural networks are insensitive to location.

To address this issue, we resort to introducing some clinical knowledge for detecting cardiac abnormalities into deep neural networks, and develop a new method, called \textbf{H}andcrafted-\textbf{R}ule-enhanced \textbf{N}eural \textbf{N}etwork (\ourapproach), to improve ECG classification accuracy. In this work, some critical information for ECG classification (e.g., the voltage of R peak, the interval of Q wave) is provided by some handcrafted rules to deal with the drawbacks of known neural networks. \ourapproach, for ECG classification with standard 12-lead ECG input, consists of a  rule inference module and a deep learning module. To combine the rule-based method and deep learning based method, \ourapproach treats both of them as meta learners, and employs a super learner to combine their predictions. 

We evaluate \ourapproach on two large-scale public ECG datasets for multi-label classification of 34 and 55 kinds of abnormalities, respectively. Comprehensive experiments demonstrate the superior performances of \ourapproach for cardiac abnormality detection in comparing with state-of-the-art methods. In a case study, the results suggest that our proposed model can be applied to detect mislabeled samples, which can help improve the annotation accuracy of ECG datasets.

The major contributions of this work are as follows:
\begin{itemize}
    \item We construct a rule-enhanced module to help promote neural networks, by providing rules according to diagnostic knowledge for ECG analysis. This design aims to provide clinical interpretation for arrhythmia with a higher consistency with experts' attention on ECG.
    \item Our model surpasses current state-of-the-art methods on two large-scale public ECG classification datasets, verifying the effectiveness of the handcrafted rules we use.
    \item In a case study, we show that our model is able to assist in detecting mislabeled samples, which is possibly beneficial to some practical tasks including corrupted label correction and AI-assisted annotation.
\end{itemize}

\section{Related Work}
Automated diagnosis has recently witnessed a rapid progress due to the fast development of deep neural networks (DNNs). In particular, many efforts have been dedicated to extending deep neural networks and designing DNN models for ECG classification.
Hannun et al.~\cite{hannun2019cardiologist} first developed
a DNN to classify 12 arrhythmia classes and demonstrated that an end-to-end DNN can classify a broad range of distinct arrhythmias from single-lead ECGs with high diagnostic performance. Many improvements were based on it. Wang et al.~\cite{wang2020weighted} added graph attention networks to capture class dependencies. Luo et al.\cite{luo2019multi} combined bi-directional \textbf{l}ong \textbf{s}hort \textbf{t}erm \textbf{m}emory (LSTM) with DNN to capture temporal features, dividing ECG data into 9 classes. However, these methods did not take into account whether a DNN correctly focuses on the key information in ECG signals to detect the corresponding classes, which might yield inferior performance in identifying some diseases.

On the other hand, many researchers attempted to utilize clinical knowledge in automated ECG analysis. Zhang et al.~\cite{zhang2014heartbeat} proposed a disease-specific feature
selection method to select ECG features and classify ECGs into five types.
% , including the ``normal'', ``supraventricular ectopic beat'', ``ventricular ectopic beat'', ``fusion of ventricular and normal beat'', and ``heartbeats that cannot be classified''. 
Xu et al.~\cite{xu2015rule} used handcrafted rules in morphological classification for ST
segments, and obtained more detailed and better results than the previous neural network methods~\cite{dalu2012detection}, suggesting possible clinical significance. Jin et al.~\cite{jin2017classification} combined a rule inference method and a convolutional neural network (CNN) to classify ECG data into normal and abnormal classes; they used some statistics (mean and variances) to depict the heart rate for diagnosis. Sannino et al.~\cite{sannino2018deep} detected peaks and waves to extract ECG temporal features, and leveraged a neural network to classify normal and abnormal ECG cases. In~\cite{mondejar2019heartbeat}, Mond{\'e}jar et al.~used \textbf{s}upport \textbf{v}ector \textbf{m}achine (SVM) to cope with temporal and morphological information (e.g., RR intervals of ECGs, wavelets, and several amplitude values). However, they simply constructed some rules or fed hand-crafted features to the models, resulting in sub-optimal performances. We argue that it would be more effective to combine rule-based features and neural networks. In addition, these methods were only presented to classify limited types of abnormalities (binary classification in most cases) and could not identify multiple diseases simultaneously, which did not meet the needs in clinical practice.

Compared with the aforementioned methods, our proposed method does not simply feed ECG features to classifiers. In contrast, we construct different rules for different abnormalities based on clinical knowledge, and fuse the rule-based outputs and multi-label predictions of a deep learning model. 

\section{Methodology}
In clinical practice, clinicians often analyze ECG signals by focusing on certain particular segments when determining whether specific abnormalities are observed.
%from two main aspects: (1) experienced doctors analyze whether the shapes of some key segments are normal, including P waves, T waves and QRS complexes; (2) clinicians quantitatively analyze voltages and duration of certain waves. 
For instance, an ECG signal is determined to be a case of ``low QRS voltage'' if the amplitudes of all the QRS complexes in the limb leads are $<0.5$ mV, and the amplitudes of all the QRS complexes in the precordial leads are $<1.0$ mV. Deep neural networks have been shown to have advantages in capturing shape patterns, leading to breakthroughs in image recognition. Also, motivated by clinical practice, we think rule-based methods can well complement deep neural networks for ECG analysis.

Based on the above motivation, we design a new handcrafted-rule-enhanced neural network (\ourapproach) for identifying ECG abnormalities (see Fig.~\ref{frame} for an overview of \ourapproach). We seek to leverage the power of a convolutional neural network (CNN) for automatic feature extraction, while employing a rule-based module to provide some key information that has been proven to be useful in clinical ECG diagnosis. To better combine these two components, we construct a super learner to fuse what the neural network learns and what the rule inference module provides. 
%In this process, 
The raw ECG signal with the gender and age information is fed to both the rule inference module and the deep learning module (see Fig.~\ref{frame}), and their outputs are fused by the proposed super learner for the final diagnosis prediction.

In what follows, we describe the deep learning module in Sec.~\ref{2A}, the rule inference module in Sec.~\ref{2B}, and the proposed super learner in Sec.~\ref{2C}. The training strategy of \ourapproach is discussed in Sec.~\ref{2D}.

\subsection{Automatic Feature Extraction by a Deep Neural Network}~\label{2A}
%In recent years, deep neural networks have shown potential in overcoming the challenges in automatic data feature extraction~\cite{gu2018recent}. 
As previous research has demonstrated that CNNs can well process ECG signals~\cite{hannun2019cardiologist,chen2020flow}, we develop an end-to-end CNN to cope with ECG signals for arrhythmia classification.

\begin{figure}[t]
\centering
\includegraphics[width=0.9\linewidth]{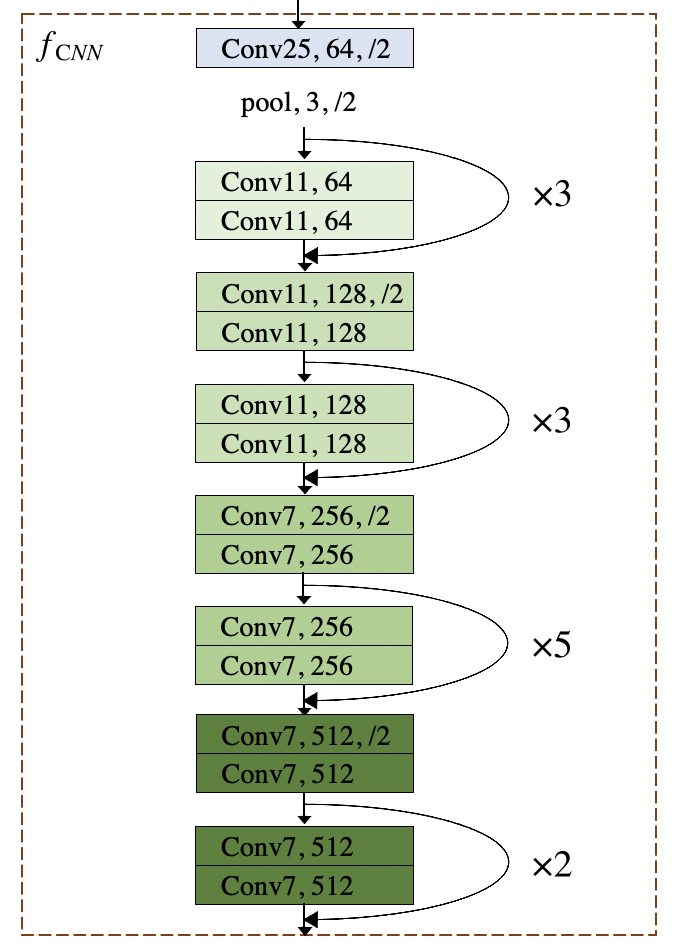}
\caption{Our $f_{\text{CNN}}$ architecture. Different from ResNet-34~\cite{he2016deep}, we use 1D convolution operation since we treat the input ECG as 1-dimensional signals. ``Conv$i, j$'' represents a convolutional layer with $j$ kernels of size $i$. If followed by ``/2'', it represents a convolutional layer with stride 2. ``Pool, 3, /2'' indicates a maxpool layer with kernel size 3 and stride 2. ``$\times n \ (n\in\{2,3,5\})$'' indicates that the residual block repeats $n$ times in sequence.}\label{fcnn}
\end{figure}

An overview of \ourapproach is shown in Fig.~\ref{frame}, in which the architecture of the CNN part (denoted by $f_{\text{CNN}}$) is specified in Fig.~\ref{fcnn}. This CNN architecture is modified from ResNet-34~\cite{he2016deep} for image classification. In Fig.~\ref{fcnn}, a ``Conv'' layer indicates a 1D convolution layer, and we use the identical mapping in the shortcut connections. The batch normalization layers and rectified linear unit (ReLU) layers are placed as in the original ResNet-34, which are not shown in Fig.~\ref{fcnn}.

The input of deep learning module includes three items: the raw ECG signal, the patient's age (encoded as a one-hot vector of size 10), and the patient's gender (a scalar). The deep learning module outputs a probability vector for possible abnormality classes. Specifically, the input ECG signal is specified as $x \in \mathbb{R}^{w_0 \times d_0}$, where $w_0$ and $d_0$ are the length of the signal and the number of leads, respectively. The output of $f_{\text{CNN}}$ is $z \in \mathbb{R}^{w \times d}$, which is further fed to a global average pooling layer for global semantics abstraction. The global average pooling is utilized to compress $z \in \mathbb{R}^{w \times d}$ into $z^\prime \in \mathbb{R}^d$. Because the patient's age and gender also affect the diagnosis results (according to clinicians' viewpoint), we concatenate the age feature vector and gender feature with $z^\prime$, and make the prediction (denoted by a vector $h \in \mathbb{R}^{N}$) for abnormality categories via a fully-connected layer with a sigmoid function (see Fig.~\ref{frame}). We define the scalar feature of gender as: (i) if the gender is missing, it is converted to 0; (2) if the gender is ``male'', it is 1; (3) if the gender is ``female'', it is 2. For the feature vector of age, we assume that the patient’s age is smaller than 100, and encode the numerical age as a 10-dimensional vector. If the age is missing, the vector is filled with zeros. In all the other cases, the age vector is a one-hot vector, where a ``1'' in the $i$-th position indicates that the patient's age is in the range of $[10 (i - 1), 10i)$.

Our design allows to modify $f_{\text{CNN}}$ or replace it by other end-to-end deep learning models. For example, it is possible to add the squeeze and excitation blocks \cite{hu2018squeeze} to the 1D convolutional layers of our proposed $f_{\text{CNN}}$ architecture. 

\subsection{Rule Inference Module}~\label{2B}
\begin{figure}[t]
\centering
\includegraphics[width=.9\linewidth]{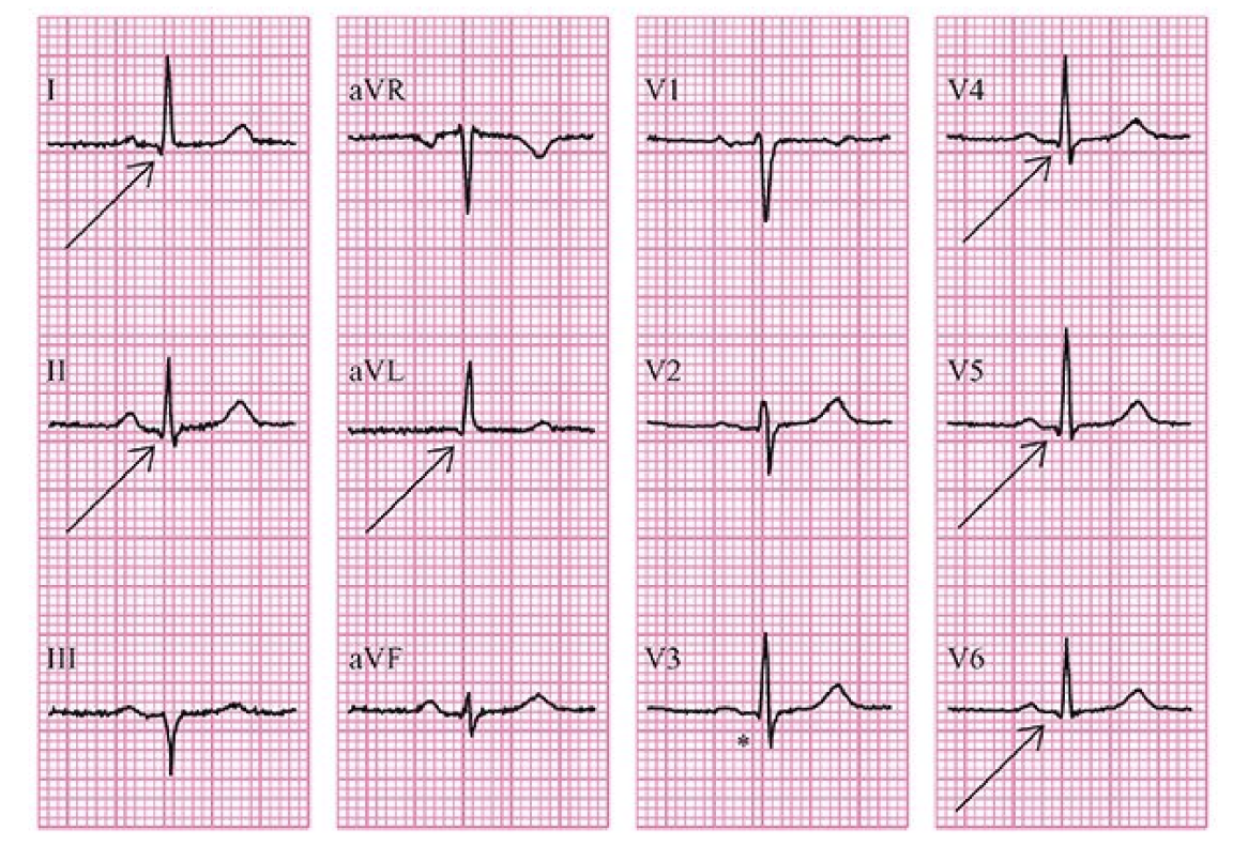}
\caption{A standard 12-lead ECG presented in the classical format used in the clinical setting \cite{Marriott0Marriott}, where one fine grid (1mm) on the vertical axis indicates 0.1mV amplitudes of the ECG signal, and the ECG recording speed is typically 25 mm (i.e., covering 2.5 fine grids along the horizon axis) per second.
}\label{classicalEcg}
\end{figure}
In clinical practice, some cardiac abnormalities can be detected by quantitative analyses, measuring, e.g., the enlargement of heart muscle, electrical conduction delay or blocks, insufficient blood flow, and death of heart muscle due to coronary thrombosis~\cite{Marriott0Marriott}.

Fig.~\ref{classicalEcg} gives an example for visualizing the classical display format of ECG signals\footnote{This example is borrowed from~\cite{Marriott0Marriott}.}.
% on its vertical axis and time on its horizon axis. 
% Measurements along the horizontal axis indicate signal and  the overall heart rate, regularity, and the time intervals during electrical activation that move throughout the heart. Measurements along the vertical axis indicate the voltage measured on the body surface.
Some abnormalities can be detected by measurements on a single ECG recording, while other abnormalities become apparent only by observing several leads and such diagnosis can be relatively complicate. As discussed in Sec.~\ref{intro}, neural networks might ignore some clinically useful information, and thus it is our desire to introduce the assistance of rule-based methods.

\begin{figure}[htbp]
\centerline{\includegraphics[width=1.\linewidth]{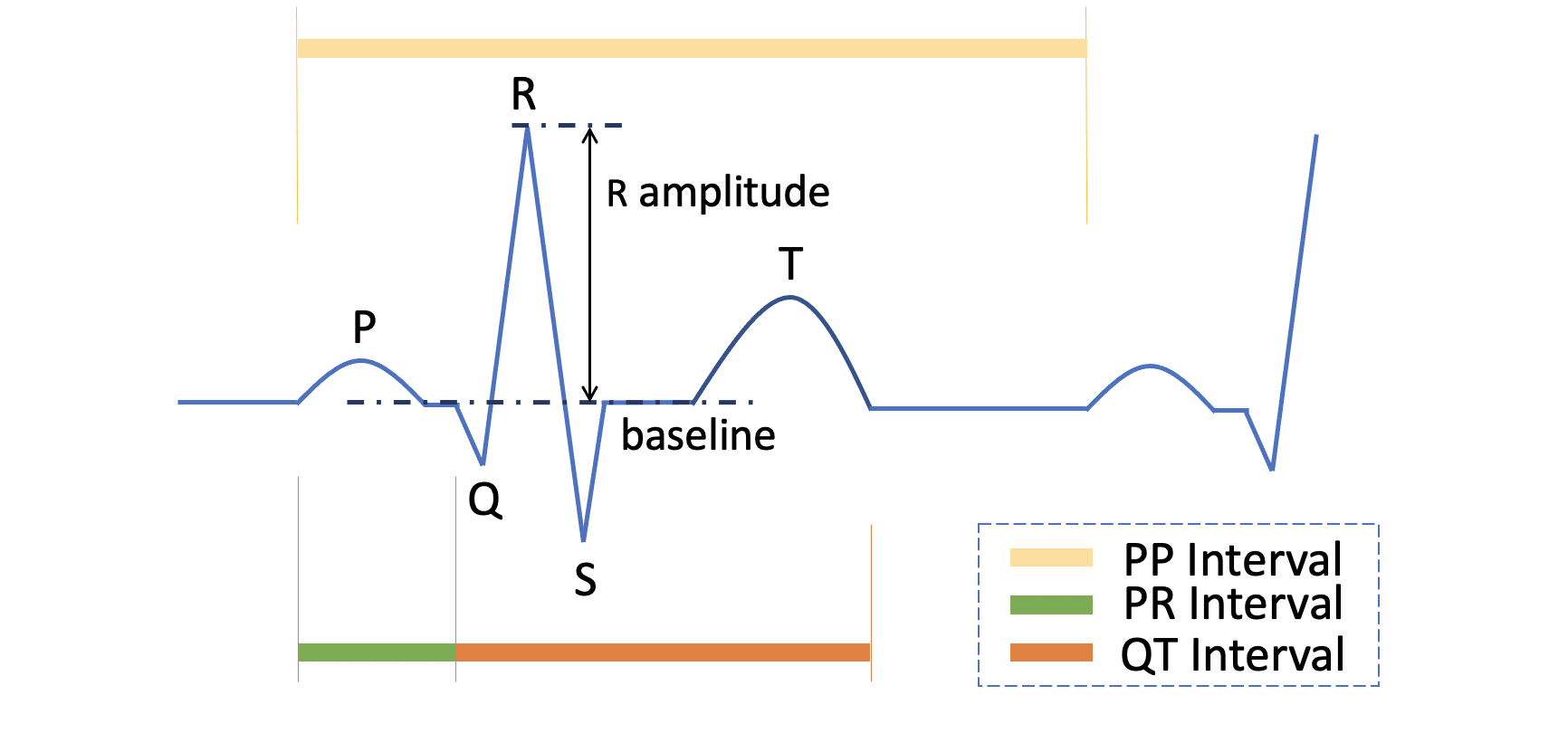}}
\caption{An example for a single cardiac cycle of an ECG signal.}
\label{ECGmorph}
\end{figure}

\begin{figure}[htbp]
\centerline{\includegraphics[width=1.\linewidth]{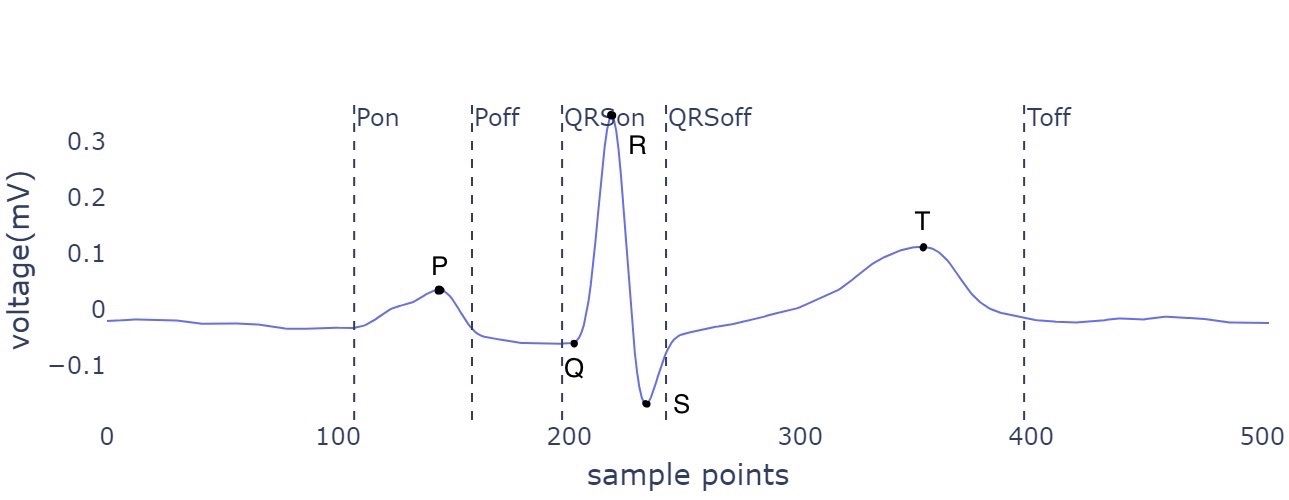}}
\caption{An example of the segmentation and delineation results on a real ECG case.}
\label{delineation}
\end{figure}

In this design, we propose a rule inference module, which performs some handcrafted rules on ECG signals to introduce certain clinical information. This rule inference module consists of a ``Segmentation and Delineation'' part and a ``Rule Inference'' part.
% and paid more attention to key segment, finally output the prediction that aligns with how clinicians make diagnosis based on ECG. 
Our ``Segmentation and Delineation'' part processes the ECG signals to obtain the cardiac cycles and key segments of ECG (see Fig.~\ref{ECGmorph}). First, we apply a band-pass (3-50Hz) filter to filter the ECG signal, so as to deburr the signal and remove the signal offsets. Second, we process R peak detection and ECG segmentation. There are methods available in BioSPPy~\cite{biosppy} to process the R peak detection and ECG segmentation. In this paper, we follow the work in~\cite{bian2021processing}, and utilize the first-order information and second-order information of the filtered ECG signal to delineate the P wave, QRS complex, and T wave. In experiments, we find that the ``Segmentation and Delineation'' part works well, and consequently, the exact cardiac cycles and key segments are obtained (see Fig.~\ref{delineation}).

After obtaining the cardiac cycles and key segments, we formulate rules for 15 ECG abnormalities according to criteria in the previous work~\cite{2015Poor,Marriott0Marriott,2006ECG}. The rules are somewhat simplified, but still maintain a high degree of consistency with experts’ attention on ECGs. 
% Our experiments show that compared with neural networks, our rule inference module is more able to find abnormalities that are more related to voltage and specific positions, and they are more interpretable. 
The cardiac abnormalities measured by these rules and the corresponding rule formulas are reported in~\autoref{rule}, in which $t(s)$ means the duration of a segment $s$, $A(s|l)$ means the amplitude of a segment $s$ in the lead $l$, $l \in \{\text{I, II, III, V1-V6, aVL, aVR, aVF}\}$, $v(s|l)$ indicates the voltage value of a segment $s$ in the lead $l$, which is presented as a vector of the same length as a segment $s$, and $v(l)$ indicates the baseline voltage value of the ECG signal in the lead $l$. In~\autoref{rule}, we compute $t(s)$ (the interval time of a segment $s$) as in Eq.~(\ref{etime}), where $len(s)$ is the number of the points recorded in the ECG signal, and $sr(s)$ is the sampling rate of the signal:
\begin{equation}
    t(s) = len(s) / sr(s) \times 1000 \text{(ms)}
\label{etime}
\end{equation}

We compute $A(s|l)$ as in Eq.~(\ref{eA}), in which $v(l)$ means the baseline amplitude of the ECG signal. Generally, clinicians regard the 50th percentile of the voltage value of a segment from Toff to Pon as the baseline amplitude. For simplicity, $v(l)$ could be computed as 0. 
\begin{equation}
\begin{aligned}
    A(s|l) = \text{max}((v(s|l) - & v(l)) \times u)\text{(mV)},\\
    & \text{if $s$ is an ``upper arch''-shaped segment} \\
           = \text{min}((v(s|l) - & v(l)) \times u)\text{(mV)},\\
    & \text{if $s$ is a ``downbend''-shaped segment}
\end{aligned}
\label{eA}
\end{equation}
where $u$ is the unit voltage defined in the dataset. One can see that the rules define some abnormalities in a precise quantification way. In the training and inference phases, the rule inference module is performed on ECG signals without back-propagation, and yields predictions for various abnormalities. If an ECG signal meets a rule formula, then the rule inference module returns ``1'' for this abnormality category, and if not, the rule inference module returns ``0''. The predictions for all the abnormalities are concatenated into a vector, and are forwarded for final prediction.

\begin{table*}[t]
\caption{The formulas we use for identifying ECG abnormalities according to criteria in the medical literature~\cite{2015Poor,Marriott0Marriott,2006ECG}. $t(s)$ indicates the duration of a segment $s$, computed according to Eq.~(\ref{etime}). $A(s|l)$ indicates the amplitude of a segment $s$ in the lead $l$, computed according to Eq.~(\ref{eA}). }
\label{rule}
\centering
\begin{threeparttable}
\begin{tabular}{ll}
\hline
\textbf{ECG Abnormalities}                       & \textbf{Rule Formulas}                                                                                                \\ \hline
\multirow{2}{*}{Poor R-wave progression\tnote{1}}              & (a)       $A(\text{R}|\text{V1}) > A(\text{R}|\text{V2}) >  A(\text{R}|\text{V3}) > A(\text{R}|\text{V4})        $                    \\
                                                      & (b)       $A(\text{R}|\text{V2})>0 \quad \& \quad A(\text{R}|\text{V3})>0 \quad \& \quad A(\text{R}|\text{V1}) + A(\text{R}|\text{V2}) + A(\text{R}|\text{V3}) <  0.2\text{mV}$ \\ \hline
Arrhythmia                                            & $\text{STD}(t(\text{PP})) > 120\text{ms}$ \tnote{2}                                                           \\ \hline
Tachycardia                                           & $\text{heart rate} > 120       $                                                                          \\ \hline
Bradycardia                                           & $\text{heart rate} < 60        $                                                                             \\ \hline
Right axis deviation (RAD)\tnote{3}                      & $-2 \times A(\text{QRS}|\text{III}) < A(\text{QRS}|\text{I}) < 0 \quad \& \quad A(\text{QRS}|\text{III}) > 0    $                            \\ \hline
Left axis deviation (LAD)                              & $A(\text{QRS}|\text{I}) > 0 \quad \& \quad A(\text{QRS}|\text{III}) < \-A(\text{QRS}|\text{I}) $                                              \\ \hline
Low QRS voltage                                       & $A(\text{QRS}|l) < 0.5\text{mV}$, $l \in \{\text{I, II, III}\} \quad or \quad A(\text{QRS}|l) < 1\text{mV}$, $l \in \{\text{V1,   V2, V3}\}  $            \\ \hline
QT prolongation                                       & $t(\text{QT})>0.4\text{s} \quad \& \quad t(\text{QT})/(\text{heart rate})^{-\nicefrac{1}{2}}>0.43$     \\ \hline
Clockwise rotation (CR)   & $0.9 <A(\text{R}|\text{V1})/A(\text{S}|\text{V1})< 1.1 \quad \& \quad 0.9 < A(\text{R}|\text{V2})/A(\text{S}|\text{V2}) < 1.1 $                \\ \hline
Counterclockwise rotation (CCR)                        & $A(\text{R}|l)/A(\text{S}|l) < 1$, $l \in \{\text{V1-V4}\}   $                                                                \\ \hline
First degree atrioventricular block                   & $t(\text{PR}) > 200\text{ms}         $                                                                           \\ \hline
Abnormal Q waves                                      & $A(\text{Q}|l) > 1/4\times A(\text{R}|l)$, $l \in \{\text{II, III, aVF}\}\quad  or\quad t(\text{Q}) > 40\text{ms}          $                   \\ \hline
T wave change (T change)                               & $A(\text{T}|l) < 1/10 \times A(\text{R}|l) \quad or\quad A(\text{T}|l) > 0.5\text{mV}$, $l \in \{\text{I, II, V2-V6}\}  $                           \\ \hline
Right atrium enlargement (RAE)                         & $A(\text{P}|l) \ge 0.15\text{mV}$, $l \in \{\text{V1,V2}\} \quad \& \quad A(\text{P}|l) \ge 0.25\text{mV}$, $l \in \{\text{II, III,   aVF}\}  $  \\ \hline
\multirow{4}{*}{Left ventricular high voltage (LVHV)\tnote{4}} & (a) $A(\text{R}|\text{V5}) > 2.5\text{mV} \quad \& \quad    A(\text{R}|\text{V6}) > 2.5\text{mV}          $                               \\
                                                      & (b) $ A(\text{R}|\text{V5}) + A(\text{S}|\text{V1}) >  4.0\text{mV} \quad \text{if} \ (\text{gender}=\text{male}) \quad \text{or}\quad >3.5\text{mV} \quad \text{if} \ (\text{gender}=\text{female})$ \\
                                                      & (c)  $   A(\text{R}|\text{I}) > 1.5\text{mV} \quad \text{or} \quad    A(\text{R}|\text{aVL}) >1.2\text{mV} \quad \text{or} \quad A(\text{R}|\text{aVF}) > 2.0\text{mV} $          \\
                                                      & (d)  $  A(\text{R}|\text{I}) + A(\text{S}|\text{III})  >  2.5\text{mV}    $                                                        \\ \hline
\end{tabular}
\begin{tablenotes}
        \footnotesize
        \item[1](a) or (b) is sufficient to detect the Poor R-wave progression.
        \item[2]STD denotes standard deviation computing.
        \item[3]$A(\text{QRS}|l) = A(\text{Q}|l) + A(\text{R}|l) + A(\text{S}|l)$
        \item[4](a) or (b) or (c) or (d) is sufficient to detect the left ventricular high voltage.
      \end{tablenotes}
    \end{threeparttable}
\end{table*}

\subsection{A Super Learner for Prediction Fusion}~\label{2C}
To combine the predictions provided by the rule inference module and the deep learning module, it is desirable to model the dependencies between these two methods and fuse their outputs for the final cardiac abnormality identification. Here we treat both the rule inference module and the deep neural network as meta learners, and introduce a super learner to fuse their predictions.

The concept of ``super learner'' was first proposed in~\cite{van2007super}, which is a weighted combination of the predictions of the meta learners. In our design, the input to the super learner is two prediction vectors produced by the deep learning module, $h_{\text{DL}} = [h_1, h_2, \ldots, h_N]$, and by the rule inference module, $l_{\text{RULE}} = [l_1, l_2, \ldots, l_N]$, where $N$ is the number of abnormalities, and the value of the $i$-th element indicates the predicted probability of the $i$-th abnormality (or normality). By fusing these two predictions, the proposed Super Learner produces the final prediction vector, $\hat{y} = [\hat{y}_1, \hat{y}_2, \ldots, \hat{y}_N]$. Formally, the operation of the Super Learner is defined in Eq.~(\ref{sl}), where $w$ is a learnable weight vector of size $N$, $s(w)$ is the sigmoid function, and ``$\cdot$'' denotes the element-wise dot multiplication:
\begin{equation}\label{sl}
    \hat{y} = h_{\text{DL}} \cdot s(w) + l_{\text{RULE}} \cdot (1-s(w))
\end{equation}

In particular, if the rule inference module considers fewer than $N$ kinds of abnormalities, a mask vector is constructed as $mask = [m^i]_{i=1}^N\ (m^i \in \{0,1\})$. If the $i$-th abnormality (or normality) 
is predicted by the rule inference module, then the $i$-th element $m^i=1$; otherwise, $m^i=0$. The learnable weight vector $w$ is of the same length as the prediction vector produced by the rule inference module, and its elements corresponding to the predictions that are not provided by the rule inference module are padded with zeros (requiring no gradients) to align with $h_{\text{DL}}$. The prediction vector $l_{\text{RULE}}$ is padded in the same way. Thus, $w$, $l_{\text{RULE}}$, and $h_{\text{DL}}$ all have the identical size of $N$.
The masked mechanism for the final prediction is performed as in Eq.~(\ref{masked}), where $\overline{mask}$ indicates to invert every element in the $mask$ vector, and ``$\cdot$'' denotes the element-wise multiplication: 
\begin{equation}
\begin{aligned}
    \hat{y} = mask \cdot (h_{\text{DL}} \cdot s(w) + l_{\text{RULE}} \cdot (1-s(w))) + \overline{mask}\cdot h_{\text{DL}}
\end{aligned}
\label{masked}
\end{equation}

\subsection{Model Training}~\label{2D}
The entire framework \ourapproach is specified using a newly proposed loss function $\mathcal{L}$ in the training phase. Since abnormal ECG cases are usually much fewer in the real world data (the datasets), the deep neural network part might fall into a pattern collapse and indiscriminately return ``zeros'' (indicating the ``normality'' category). Thus, in designing the loss function, we manage to use the predictions generated by the rule inference module to guide the predictions of the deep learning module, because the rule inference module often performs better on the categories with fewer cases in the datasets.

Assume that the ground truth of an ECG record is $y = \{y^i\}_{i=1}^N, y^i \in \{0,1\}$. When $y^i=1$, it means that the ECG record is with the abnormality of category $i$, and $y^i=0$ is for normality of category $i$. The weighted cross-entropy loss is defined by:
\begin{equation}
\mathit{L}(y, \hat{y}) = \sum_{i=1}^N w_i \cdot y^i \log \hat{y}^i + (1-y^i)\log(1 - \hat{y}^i)
\label{multie}
\end{equation}
where $w_i$ is the class weight of category $i$ computed as $w_i = \frac{M}{M_i}$ ($M$ is the total number of all ECG cases in the dataset and $M_i$ is the number of cases belonging to category $i$), and $\hat{y}$ is the prediction probabilities obtained by the Super Learner. The predictions of the rule inference module are also used to guide the prediction of \ourapproach, and its total loss function is defined by:
\begin{equation}
\mathcal{L} = \mathit{L}(y,\hat{y}) + \lambda \mathit{L}(l_{\text{RULE}},\hat{y})
\label{loss}
\end{equation}
where $\mathit{L}$ is the weighted binary cross-entropy in Eq.~(\ref{multie}), $l_{\text{RULE}}$ is the output probabilities generated by the rule inference module, and $\lambda$ is an importance parameter.

\section{Experiments}
\subsection{Datasets}
\begin{figure}[htbp]
\centerline{\includegraphics[width=1\linewidth]{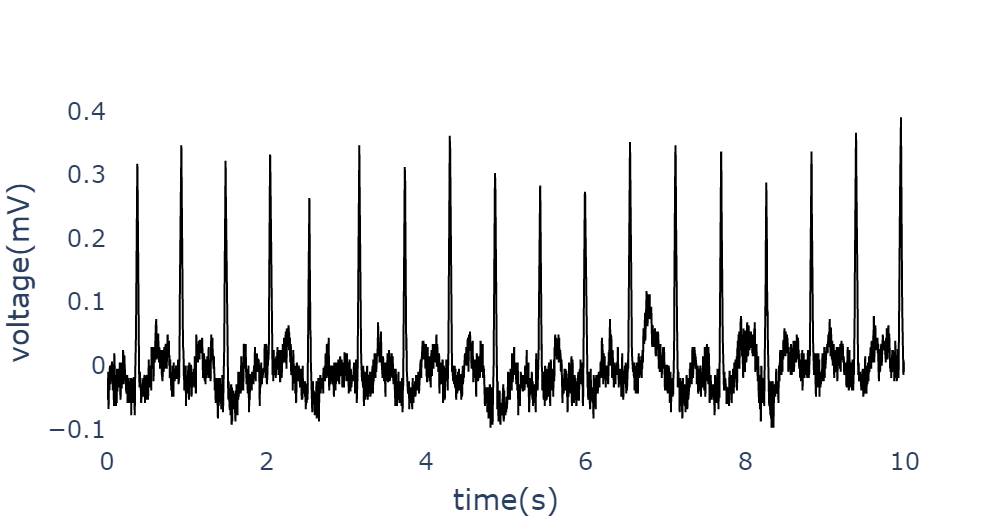}}
\centerline{\includegraphics[width=1\linewidth]{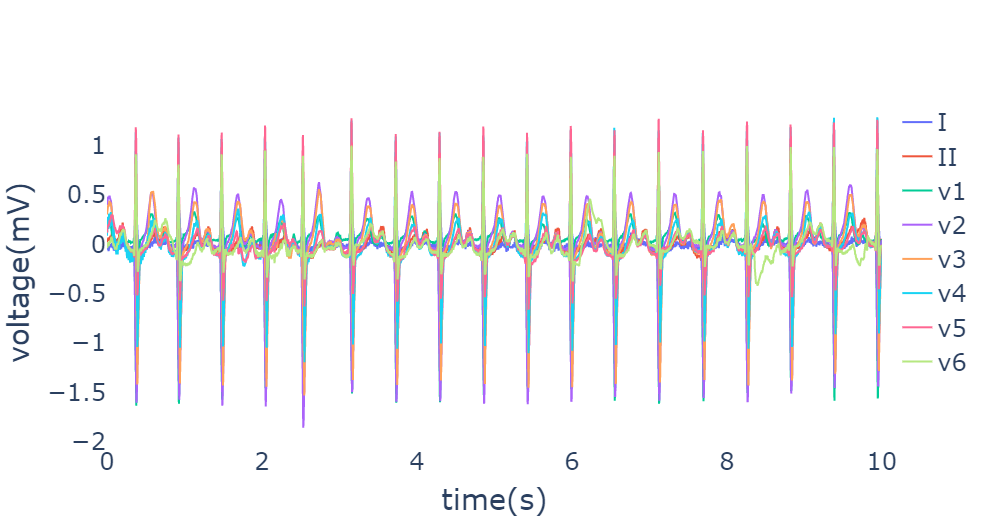}}
\caption{An ECG signal example with lead I (top), and an ECG signal example with 8 leads (bottom).}
\label{plotsample}
\end{figure}

We use the datasets for the first and second rounds of the contest of the Hefei Hi-tech Cup ECG Intelligence Competition\footnote{\url{https://tianchi.aliyun.com/competition/entrance/231754/information}} for a multi-label classification task of 55 classes and 34 classes, respectively. In clinical ECG diagnosis, clinicians often give detailed analysis with the pathogenesis and types of abnormalities. The datasets we use cover comprehensive ECG disease classes that are commonly found in clinical application scenarios. The dataset used in the first round of the contest (called ``TianChi ECG dataset-1'') contains 24,106 samples, and the second round dataset (called ``TianChi ECG dataset-2'') contains 20,096 samples. Each sample has 8 leads (I, II, V1, V2, V3, V4, V5, V6). Each sample was recorded in 10 seconds with 500 Hz sampling frequency, and the unit voltage is $4.88\times10^{-3}$ millivolts. For these two datasets, the ECG input for $f_\text{CNN}$ is with a shape of $5000 \times 12$. Since the standard 12-lead ECG is the most commonly-used format in ECG analyses~\cite{0Automatic}, we computationally added the other four leads to the original datasets following Eq.~(\ref{lead}). An example of the input ECG is shown in Fig.~\ref{plotsample}.

\begin{equation}\label{lead}
\begin{aligned}
&v(\text{III}) = v(\text{II}) - v(\text{I})  \\
&v(\text{aVR}) = -(v(\text{I}) + v(\text{II}))/2 \\
&v(\text{aVL}) = v(\text{I}) - v(\text{II})/2 \\
&v(\text{aVF}) = v(\text{II}) - v(\text{I})/2 
\end{aligned}
\end{equation}

The 55 cardiac abnormalities (or normalities) of TianChi ECG dataset-1 and the 34 cardiac abnormalities (or normalities) of TianChi ECG dataset-2 are shown in \autoref{table-cate}. 
%As the ECG abnormalities normally co-occur in the real world, we construct matrix $P$ from the dataset to display label dependencies, where $P(i|j)$ means when abnormality $i$ appears, abnormality $j$ has the possibility $P(i|j), 1 \ge P(i|j) \ge 0$ to appear, and $P(i|j)$ may not be equal to $P(j|i)$ in the real world. The co-occurrence matrix is in Fig.~\ref{classes}.

% \begin{figure}[htbp]
% \centerline{\includegraphics[width=.9\linewidth]{image/classes.png}}
% \caption{A adjacency matrix heat map to display the dependencies of different classes.}
% \label{classes}
% \end{figure}

\begin{table}[tb]
\centering
\caption{The record number and proportion of each class in the two Tianchi ECG datasets. Some of the categories use abbreviations. The comparison table of full names and abbreviations is given in Appendix.}
\label{table-cate}
\resizebox{\columnwidth}{!}{
\begin{tabular}{l|l|l} %{p{0.6\columnwidth}p}
\hline
\multicolumn{1}{c|}{\multirow{2}{*}{\textbf{Categories}}} &  \multicolumn{2}{c}{\textbf{\# of cases (proportion \%)}} \\ \cline{2-3}
& ECG dataset-1 & ECG dataset-2\\ \hline
Sinus rhythm                                        &16918 (70.18\%) & 9536 (47.45\%)             \\ \hline
Bradycardia                                              &3372 (13.99\%) & 5272 (26.23\%)             \\ \hline
Tachycardia                                              &2010 (8.34\%) & 4910 (24.433\%)             \\ \hline
T change                                 &3421 (14.19\%) & 3490 (17.37\%)             \\ \hline
RAD                               &1055 (4.38\%) & 1131 (5.628\%)              \\ \hline
LAD                                &1137 (4.72\%) & 1128 (5.613\%)              \\ \hline
Arrhythmia                                               &924 (3.83\%) & 904 (4.498\%)              \\ \hline
VPB                        &971 (4.03\%) & 571 (2.841\%)               \\ \hline
RBBB                         &392 (1.63\%) & 556 (2.767\%)               \\ \hline
CRBBB               &1109 (4.60\%) & 423 (2.105\%)               \\ \hline
LVHV                    &4326 (17.95\%) & 415 (2.065\%)               \\ \hline
APB                              &1470 (6.10\%) & 316 (1.572\%)               \\ \hline
ST-T change                                              &2111 (8.76\%) & 299 (1.488\%)               \\ \hline
ST change                                                &2967 (12.31\%) & 287 (1.43\%)                \\ \hline
IAVB                    &282 (1.17\%) & 142 (0.707\%)               \\ \hline
IRBBB            &199 (0.83\%) & 126 (0.627\%)               \\ \hline
Atrial fibrillation                               &1217 (5.05\%) & 120 (0.597\%)               \\ \hline
NS-ST            &78 (0.32\%) & 64 (0.318\%)                \\ \hline
CCR                          &34 (0.14\%) & 61 (0.304\%)                \\ \hline
Abnormal Q waves                                         &53 (0.22\%) & 52 (0.259\%)                \\ \hline
LABB                       &106 (0.44\%) & 36 (0.179\%)                \\ \hline
NS-T                &125 (0.52\%) & 35 (0.174\%)                \\ \hline
CR                                  &30 (0.12\%) & 35 (0.174\%)                \\ \hline
RAE                         &24 (0.10\%) & 32 (0.159\%)                \\ \hline
RVR                            &229 (0.95) & 29 (0.144\%)                \\ \hline
LBBB                         &18 (0.08\%) & 27 (0.134\%)                \\ \hline
CLBBB              &20 (0.08\%) & 27 (0.134\%)                \\ \hline
Short PR interval                                 &55 (0.23\%) & 24 (0.119\%)                \\ \hline
Early repolarization                                     &37 (0.15\%) & 22 (0.109\%)                 \\ \hline
Pacing rhythm                                            &74 (0.31\%) & 16 (0.08\%)                 \\ \hline
Poor R-wave progression                                  &19 (0.08\%) & 16 (0.08\%)                 \\ \hline
NS-STT  &61 (0.25\%) & 16 (0.08\%)                 \\ \hline
Fusion wave                                              &18 (0.08\%) & 8 (0.04\%)                  \\ \hline
QRS low voltage                                          &1543 (6.40\%) & 3 (0.015\%)                 \\ \hline
Normal ECG & 4171 (17.30\%) & -- \\ \hline
Critical ECG & 1911 (7.93\%) & -- \\ \hline
Abnormal ECG &  1061 (4.40\%)            & -- \\ \hline
LVH & 432 (1.79\%) & -- \\ \hline
QT prolongation & 101 (0.42\%) & -- \\ \hline
Differential conduction & 75 (0.31\%) & -- \\ \hline
Atrial fibrillation & 42 (0.17\%) & -- \\ \hline
Intraventricular aberrant conduction & 36 (0.15\%) & -- \\ \hline
Bigeminy & 27 (0.11\%) & -- \\ \hline
Ventricular premature beats & 33 (0.14\%) & -- \\ \hline
Abnormal repolarization & 29 (0.12\%) & -- \\ \hline
uAPB & 25 (0.10\%) & -- \\ \hline
Cor Pulmonale & 27 (0.11\%) & -- \\ \hline
SVR  & 30 (0.12\%) & -- \\ \hline
Short series of atrial tachycardia & 21 (0.09\%) & -- \\ \hline
RVE & 18 (0.08\%) & -- \\ \hline
Atrioventricular conduction delay & 10 (0.04\%) & -- \\ \hline
Bifascicular block & 20 (0.08\%) & -- \\ \hline
NS-IB & 17 (0.07\%) & -- \\ \hline
NS-ID & 16 (0.07\%) & -- \\ \hline
P pulmonale & 17 (0.07\%) & -- \\ \hline
\end{tabular}
}
\end{table}

\subsection{Experimental Setting}
We use the Adam optimizer~\cite{kingma2014adam} with default hyper-parameters. The base learning rate (base\_lr) was initialized as $10^{-4}$, and we gradually warm-up~\cite{warmup} the first 2 epochs. The learning rate scheduler is set as the cosine scheduler with a weight decay of $10^{-6}$. The batch size is 32, and the number of epochs is 60.

We conduct a comprehensive evaluation, and compare our approach with four state-of-the-art ECG signal classification models, including 1D Transformer-XL~\cite{dai2019transformer}, SE-ECGNet\cite{9313548}, 1D ResNet-34\cite{brito2019electrocardiogram}, and MLWGAT\cite{wang2020weighted}, on the two TianChi ECG datasets. We adopt the same data pre-processing method and configurations for all these methods.

We develop all the models using PyTorch, and run all the experiments on an NVIDIA GTX 2080Ti 64GB GPU machine.

\subsection{Evaluation Metrics}
Following conventional settings~\cite{wang2020weighted,hannun2019cardiologist} and taking patients' concern into account, we report the average per-class recall (CR), average per-class F1 (CF1), average overall recall (OR), and average overall F1 (OF1) for performance evaluation. For each ECG sample, the labels are predicted as positive if the probabilities for them are larger than 0.5. Generally, the average overall recall (OR) and average per-class recall (CR) are relatively more important for ECG abnormality detection in the clinical setting since neglecting a disease is much more harmful for the patient.

\subsection{Experimental Results}
%In this part, 
We present comparison results with the state-of-the-art methods. In addition, we provide some running examples for showing that our model can assist detecting mislabelled samples.

\subsubsection{Comparison with State-of-the-art Methods}
The results of our comparative evaluation experiments are summarized in \autoref{comparison1} and \autoref{comparison2}.
We report the OF1, CF1, OR, and CR of the four known models evaluated on the two TianChi ECG datasets.

\begin{table}[t]
\centering
\caption{Comparison with the four state-of-the-art methods on TianChi ECG Dataset-1.}
\label{comparison1}
\begin{tabular}{lllll}
\hline
\textbf{Method}                    & \textbf{OF1}               & \textbf{CF1} & \textbf{OR} & \textbf{CR} \\ \hline
1D Transformer \cite{dai2019transformer} &  0.7731     &     0.2814  &  0.7185  & 0.2554 \\
1D ResNet-34  \cite{brito2019electrocardiogram} & 0.8679 & 0.4988  &  0.8632   & 0.5030\\
SE-ECGNet   \cite{9313548}                 & 0.8651     & 0.5024       & 0.8545  & 0.4680     \\

MLWGAT\cite{wang2020weighted}      & 0.8401  & 0.4259    & 0.8267      & 0.3959      \\ \hline
\ourapproach$^1$      & \textbf{0.8691} & 0.4855      & \textbf{0.8653}      & 0.4852     \\
\ourapproach$^2$       & 0.6312 & \textbf{0.5071}      & \textbf{0.9065}   & \textbf{0.6318}      \\ \hline
\end{tabular}\\
\begin{tablenotes}
 \item[1] $^1$With $\lambda$ in Eq.~(\ref{loss}) set to 0.
 \item[2] $^2$With $\lambda$ in Eq.~(\ref{loss}) set to 1.
\end{tablenotes}
\end{table}

\begin{table}[t]
\centering
\caption{Comparison with the four state-of-the-art methods on TianChi ECG Dataset-2.}
\label{comparison2}
\begin{tabular}{lllll}
\hline
\textbf{Method}                    & \textbf{OF1}               & \textbf{CF1} & \textbf{OR} & \textbf{CR} \\ \hline
1D Transformer \cite{dai2019transformer} &  0.8950     &     0.3123  &  0.8601  & 0.2773 \\
1D ResNet-34  \cite{brito2019electrocardiogram} & 0.9038   & 0.4686  & 0.8872 & 0.4598 \\
SE-ECGNet   \cite{9313548}                 & 0.9019     & 0.4780       & 0.8940      & 0.4620      \\

MLWGAT\cite{wang2020weighted}                             & 0.9069                     & 0.4883       & 0.8920      & 0.4774      \\ \hline
\ourapproach$^1$      & \textbf{0.9104} & 0.4655       & \textbf{0.9018}      & 0.4619      \\
\ourapproach$^2$       & 0.7224 & \textbf{0.5001}       & \textbf{0.9520}      & \textbf{0.6402}      \\ \hline
\end{tabular}\\
\begin{tablenotes}
 \item[1] $^1$With $\lambda$ in Eq.~(\ref{loss}) set to 0.
 \item[2] $^2$With $\lambda$ in Eq.~(\ref{loss}) set to 1.
\end{tablenotes}
\end{table}

\begin{figure*}[htbp]
\centerline{\includegraphics[width=1\linewidth]{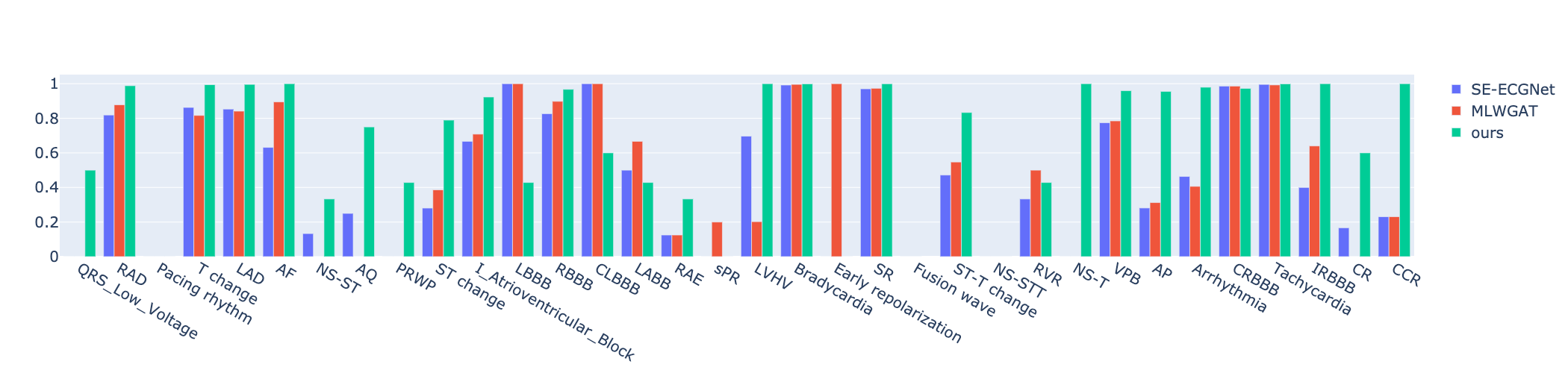}}
\caption{Comparison of sensitivities among our proposed method and the four known state-of-the-art methods. The $x$-axis indicates the 34 cardiac abnormalities (or normalities) defined in TianChi ECG dataset-2, and the $y$-axis indicates the sensitivity (the recall score).}
\label{sensitivity}
\end{figure*}

In \autoref{comparison1} and \autoref{comparison2}, we highlight the best result for each metric in bold. One can observe from the results that our model achieves the best performance on CF1, OR, and CR. 
For TianChi ECG dataset-1, with $\lambda = 0$, \ourapproach achieves the highest OF1 score and overall recall score among all the five methods. With $\lambda = 1$, \ourapproach achieves a 90.7\% overall recall score and a 63.2\% average per-class recall score, outperforming the state-of-the-art performance by over 4\% and 13\%, respectively. Meanwhile, \ourapproach achieves a 50.7\% average per-class F1 (CF1) score.
For TianChi ECG dataset-2, with $\lambda = 0$, \ourapproach achieves the highest OF1 score and overall recall score among all the five methods. With $\lambda = 1$, \ourapproach achieves a 95.2\% overall recall score and a 64.0\% average per-class recall score, outperforming the state-of-the-art performance by over 5\% and 16\%, respectively. Meanwhile, \ourapproach achieves a 50.0\% average per-class F1 (CF1) score and outperforms the state-of-the-art performance by 1.2\%. 
Somehow, with $\lambda = 1$, our overall F1 (OF1) scores are lower than the four models. This is probably mainly due to the presence of more false-positive (FP) samples.  False positive is a result indicating that a given condition occurs when it actually does not. If we detect FP samples, it means that our model outputs 1 for a specific disease but the ground truth of this disease is 0. However, we will show below that our identification of some of the FP samples is correct, and hence our OF1 score could be better in real settings. Also, considering the OF1 score and CF1 score at the same time, our model has a comparable or even better ability to perform multi-label classification, thus this decrease in OF1 score could be acceptable. 

We also display the Recall Score (sensitivity) for each class of Tianchi ECG dataset-2 in Fig.~\ref{sensitivity}. \ourapproach yields relatively high sensitivity in identifying different ECG abnormalities compared to the two state-of-the art methods. In addition, we can see that there are some categories which are not included in our rule inference module, and their sensitivities still get improved (e.g., AF, NS-ST, AQ, ST-change, etc). It might be because our model gives confidence in the rule inference module, and then back-propagation pushes the deep learning module to pay more attention to the performance of the other categories. This evidence shows that our fusion of the rule inference and deep learning network outputs is effective.

Notably, we observe that 1D ResNet-34 can also attain competitive performances on TianChi ECG Dataset-1 (see Table~\ref{comparison1}), but as Fig.~\ref{fig:gradcam} shows, the model cannot focus on the segment associated with the corresponding ECG abnormality. This might be because the deep learning model is effective for the ECG cases with obvious abnormalities, but for those abnormalities that are inconspicuous (e.g., those abnormalities that can only be observed with partial voltages and intervals), the deep learning model fails to capture accurate information. This phenomenon suggests that incorporating clinical knowledge into deep learning may provide a good potential to obtain performance improvements.
% Thus, we encourage to incorporate clinical knowledge into deep learning models still has huge potential improvements.}

\subsubsection{Ablation Studies}
We conduct experiments to verify the effect of introducing age and gender information to the neural network. We also conduct experiments to verify the contributions of the rule inference module, where we remove the meta-learner of the ``Rule Inference'' on the basis of \ourapproach and find out how the model performs. \autoref{ablation1} and \autoref{ablation2} show that the information of age and gender benefits our model, which is consistent with the previous medical observations. For instance, human's heart rate decreases with age, and the left atrial hypertrophy is also observed to be related to gender. The results in \autoref{ablation1} and \autoref{ablation2} also show that our rule inference module facilitates ECG abnormality identification with a clear margin (shown by the indicators OR and CR scores). 

In addition, we observe that equipping the rule inference module is almost free of cost for inference. Hence, it is promising to equip other neural networks with the rule inference module, which can improve performances at low cost.

\begin{table}[t]
\centering
\caption{Ablation study on Tianchi ECG Dataset-1.}
\label{ablation1}
\begin{tabular}{lllll}
\hline
\textbf{Method} & \textbf{OF1} & \textbf{CF1} & \textbf{OR} & \textbf{CR} \\ \hline
HRNN.dl$^1$ (original) & 0.8661& 0.4973& 0.8579& 0.4853  \\ \hline
HRNN.dl$^2$ &\textbf{0.8741}&0.5062& 0.8743 & 0.4854\\ \hline
HRNN ($\lambda=0$)  & 0.8691 & 0.4855      & 0.8653     & 0.4852   \\ 
HRNN ($\lambda=1$)  & 0.6312 & \textbf{0.5071}      & \textbf{0.9065}   & \textbf{0.6318} \\ \hline
\end{tabular}\\
\begin{tablenotes}
 \item[1] $^1$Only deep learning module in \ourapproach without coding information \\of gender and age.
 \item[2] $^2$Only deep learning module in \ourapproach with coding information of \\gender and age.
\end{tablenotes}
\end{table}

\begin{table}[t]
\centering
\caption{Ablation study on Tianchi ECG Dataset-2.}
\label{ablation2}
\begin{tabular}{lllll}
\hline
\textbf{Method} & \textbf{OF1} & \textbf{CF1} & \textbf{OR} & \textbf{CR} \\ \hline
HRNN.dl$^1$ (original) &0.9033& 0.4667& 0.8890& 0.4581 \\ \hline
HRNN.dl$^2$& \textbf{0.9132} & 0.4718& 0.8981& 0.4537  \\ \hline
HRNN ($\lambda=0$)  & 0.9104 & 0.4655       & \textbf{0.9018}      & 0.4619  \\ 
HRNN ($\lambda=1$)  & 0.7224 & \textbf{0.5001}       & \textbf{0.9520}      & \textbf{0.6402}  \\ \hline
\end{tabular}\\
\begin{tablenotes}
 \item[1] $^1$Only deep learning module in \ourapproach without coding information \\of gender and age.
 \item[2] $^2$Only deep learning module in \ourapproach with coding information of \\gender and age.
\end{tablenotes}
\end{table}

\subsubsection{Detection of Mislabelled Samples}
Our rule inference module is able to detect mislabelled samples, as reviewed and verified by senior certified ECG clinicians. In the following examples, dash lines delineate P wave in the ECG signals, dot lines for QRS complex, and dash-dot lines for T wave. We also add blue to the signals to illustrate the sections focused by senior certified ECG clinicians. 

\begin{figure}[htbp]
    \centerline{\includegraphics[width=1.\linewidth]{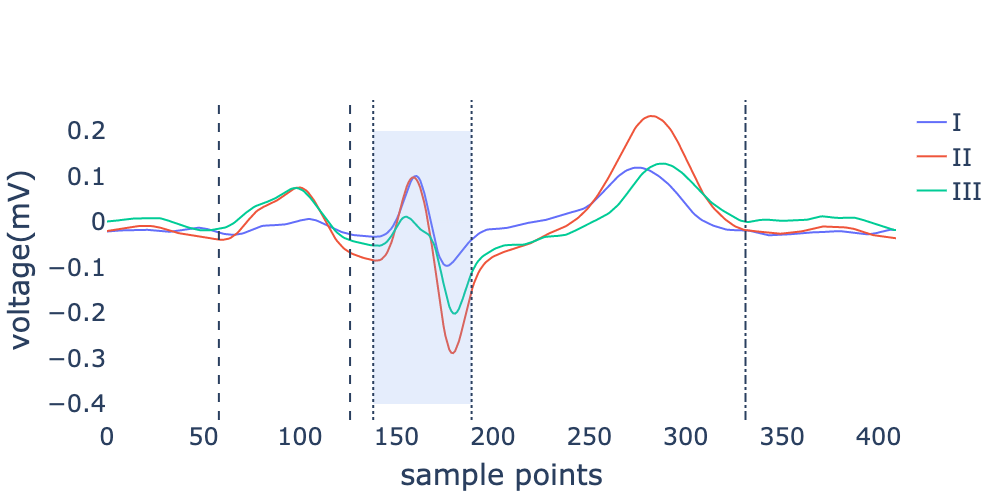}}
    \caption{Illustrating a cardiac abnormality, Low QRS Voltage, which is mislabelled as non Low QRS Voltage in the dataset.}
    \label{QRSlowvoltage}
    \end{figure}

\begin{figure}[htbp]
    \centerline{\includegraphics[width=1.\linewidth]{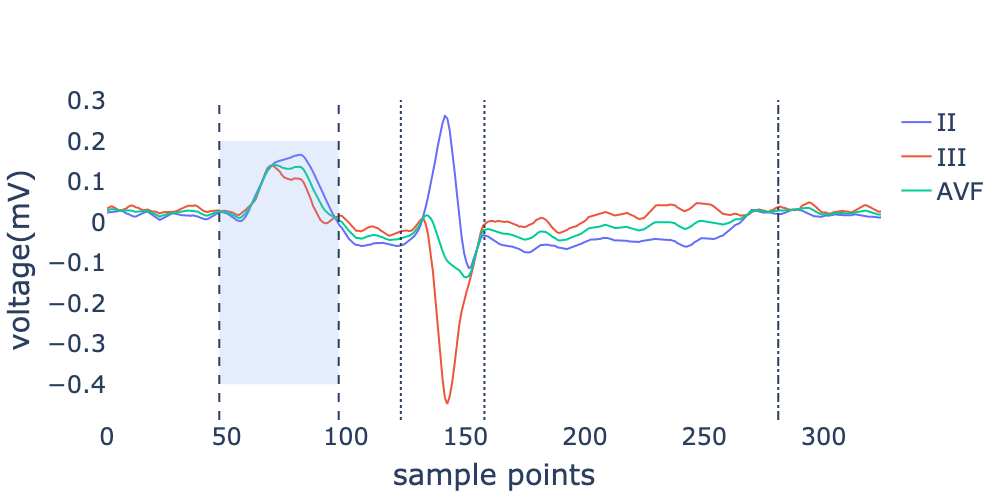}}
    \centerline{\includegraphics[width=1.\linewidth]{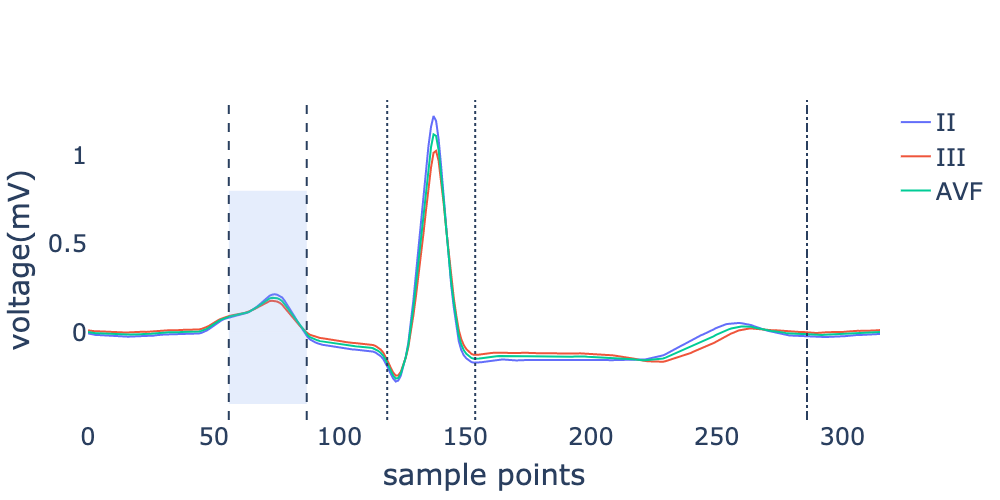}}
    \caption{An example labelled as Right atrium enlargement, which however does not have that abnormality (top); an example of RAE, mislabelled as non-RAE (bottom).}
    \label{RAE}
    \end{figure}

\begin{figure}[htbp]
    \centerline{\includegraphics[width=1.\linewidth]{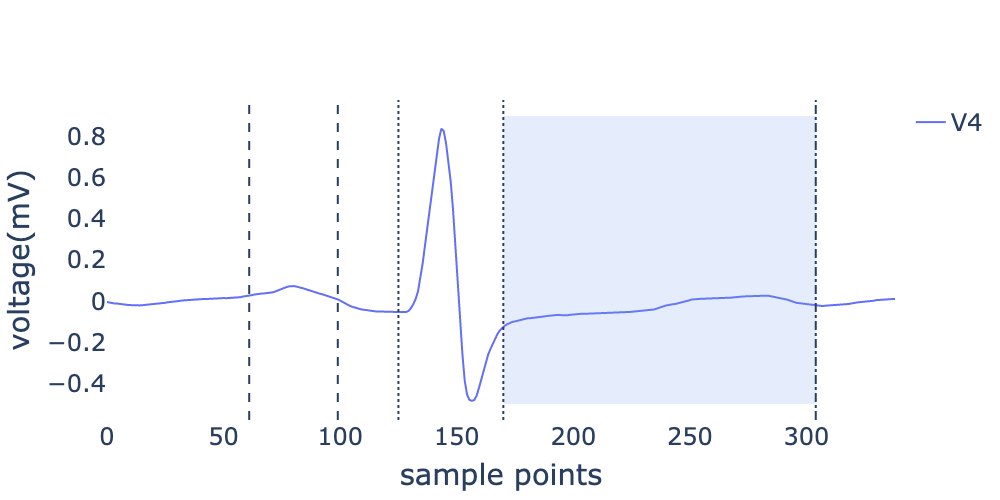}}
    \caption{An example of T wave change, mislabelled as non-T-change.}
    \label{Tchange}
    \end{figure}

\begin{figure}[htbp]
    \centerline{\includegraphics[width=1.\linewidth]{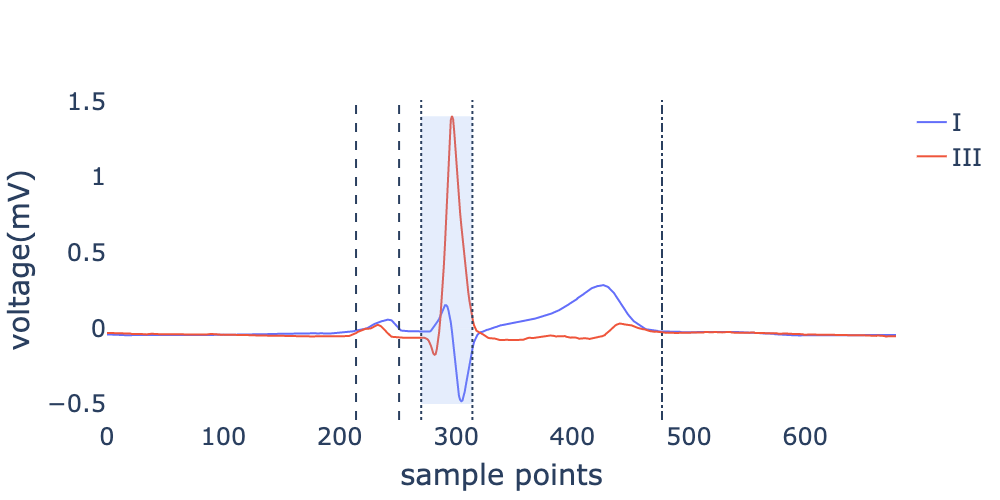}}
    \caption{An example of RAD, mislabelled as non-RAD. }
    \label{RAD}
    \end{figure}

\begin{figure}[htbp]
    \centerline{\includegraphics[width=1.\linewidth]{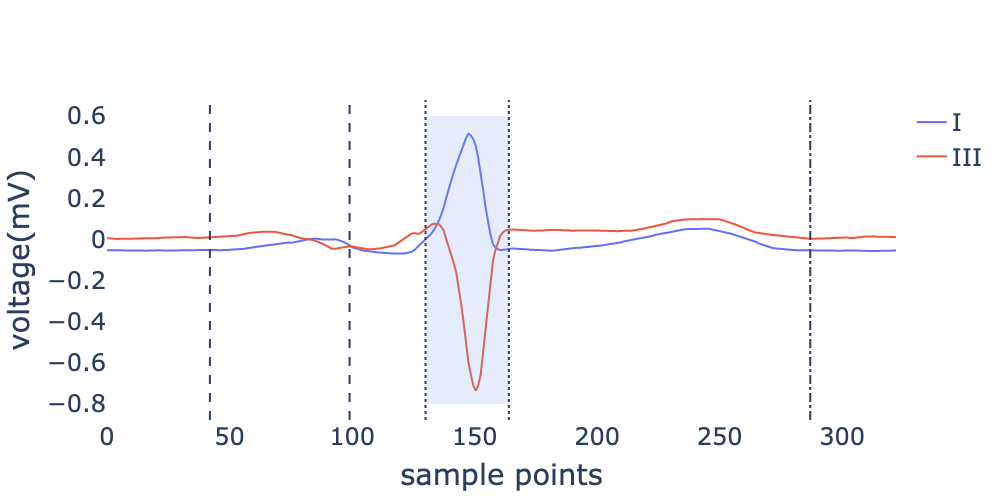}}
    \caption{An example of LAD, mislabelled as non-LAD. }
    \label{LAD}
    \end{figure}

\begin{figure}[htbp]
    \centerline{\includegraphics[width=1.\linewidth]{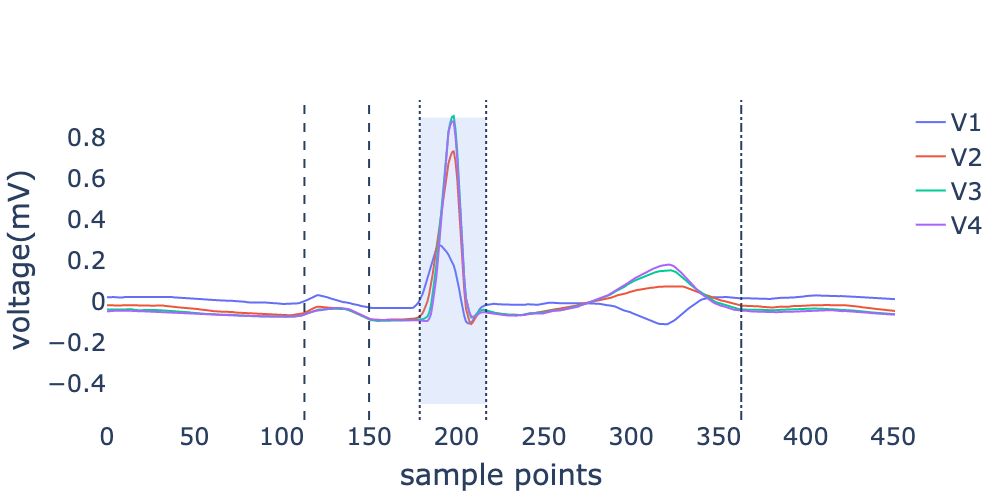}}
    \caption{An example of CCR, mislabelled as non-CCR. }
    \label{CCR}
    \end{figure}

\begin{itemize}
    \item Low QRS voltage: In Fig.~\ref{QRSlowvoltage}, we can see that QRS amplitude is less than 5mV in limb leads, which meets the requirement according to \autoref{rule}.
    \item Right atrium enlargement (RAE): In Fig.~\ref{RAE}, Right atrium enlargement produces a peaked P wave (P pulmonale) with amplitude $>$ 0.25mV in the inferior leads (II, III, and AVF). However, the sample in Fig.~\ref{RAE} does not have a peaked P wave.
    \item T wave change (T change):  In Fig.~\ref{Tchange}, the amplitude of the T wave is less than 1/10 of the R peak in the wave in the V1 lead.
    \item Right axis deviation (RAD): The sample shown in Fig.~\ref{RAD} meets the requirement of the corresponding rule in \autoref{rule}. 
    \item Left axis deviation (LAD): The sample shown in Fig.~\ref{LAD} meets the requirement of the rule for Left axis deviation in \autoref{rule}. 
    \item Counterclockwise rotation (CCR): In Fig.~\ref{CCR}, the R wave of the V5 or V6 lead appears on the V2, V3, and V4 leads; this sample could be diagnosed as CCR.

\end{itemize}


\section{Conclusions and Future Work}
In this paper, we proposed a new Handcrafted-Rule-enhanced Neural Network, \ourapproach, to classify arrhythmia diseases using ECG signals. To our best knowledge, this is the first approach that combines handcrafted rule methods and neural networks for multi-label ECG classification. 
%Experiments showed that \ourapproach is able to help identify mislabeled samples, showing potential for some practical tasks including corrupted label correction and AI-assisted annotation.
% Generally, average overall recall (OR) and average per-class recall (CR) are relatively more important for the ECG abnormality detection in the clinical setting, since ignoring a disease is much more harmful for the patient. However, due to the existence of unbalanced data sets, we find that the neural networks will be more inclined to output zero in order to minimize the loss function, which is to increase the probabilities of missed detection in the process of medical diagnosis. 
Experiments showed that \ourapproach achieves the highest overall recall score and the highest average per-class recall score on both the two TianChi ECG datasets, and outperforms state-of-the-art methods with clear margins (e.g., over 4\% in the overall recall score and 13\% in the average per-class recall score).
Meanwhile, our model attains the highest average per-class F1 score among state-of-the-art methods.
Experiments also demonstrated that \ourapproach is able to help identify mislabelled samples, showing good potential for some practical tasks including corrupted label correction and AI-assisted annotation.

There are several possible improvements and extensions to \ourapproach that we would like to pursue in future work: (1) encoding rules into neural networks and training rule modules with back-propagation; (2) leveraging graph neural networks to model the dependencies among different abnormalities and utilizing such dependencies in automatic ECG diagnosis.
% A particularly interesting research direction would be like work~\cite{hou2021coordinate} and meanwhile if we could modify the layer to help neural network more accurately capture the information of the corresponding waves like experts, the application of artificial intelligence in ECG diagnose in real clinical settings would become more prevalent. Our future work will be focused on embedding location information of ECG signals into deep learning neural networks and validation in a clinical setting.

\section*{Acknowledgment}

This research was partially supported by National Key R\&D Program of China under grant No. 2019YFB1404802, National Natural Science Foundation of China under grants No. 62176231 and 62106218, Zhejiang public welfare technology research project under grant No. LGF20F020013, Wenzhou Bureau of Science and Technology of China under grant  No. Y2020082. D. Z. Chen’s research was supported in part by NSF Grant CCF-1617735.

\ifCLASSOPTIONcaptionsoff
  \newpage
\fi



% \bibliographystyle{IEEEtran}


\bibliographystyle{IEEEtran}
\bibliography{reference}

\begin{IEEEbiography}[{\includegraphics[width=1in,height=1.25in,clip,keepaspectratio]{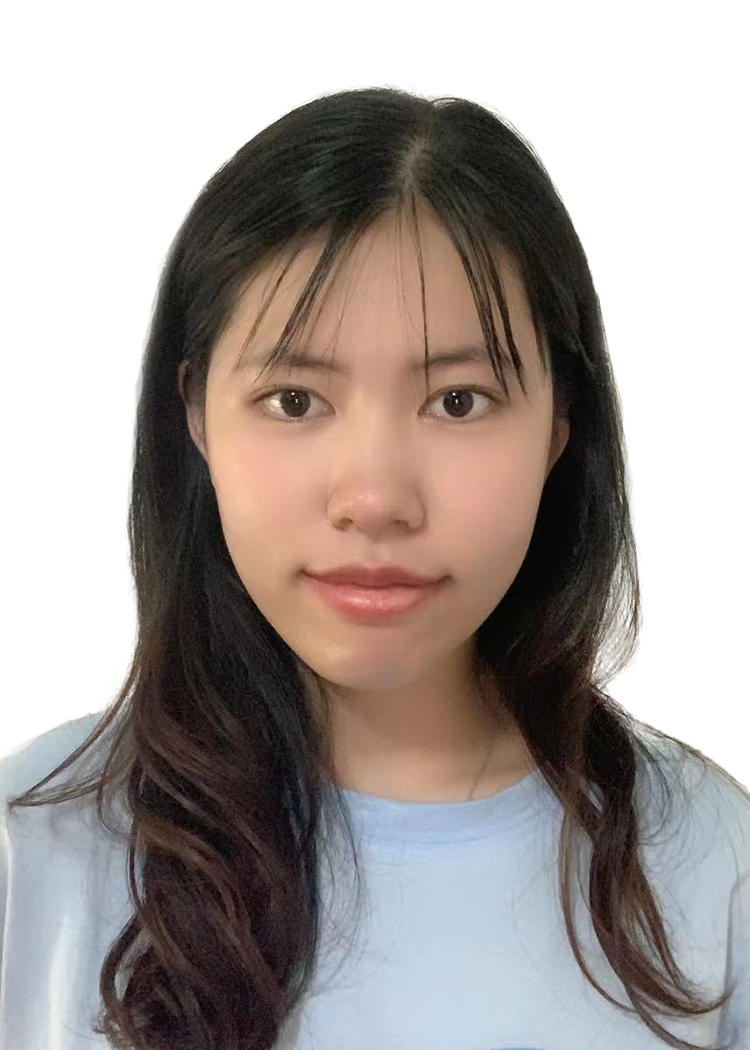}}]{Yuexin Bian}
received her B.S. degree from Zhejiang University, China, in 2021. She is now a Ph.D. student in the Department of Electrical and Computer Engineering at University of California San Diego, USA. Her research interests include machine learning, optimization for control, data mining and signal processing and analysis. Yuexin Bian
contributes to this work in Zhejiang University.
\end{IEEEbiography}

\begin{IEEEbiography}[{\includegraphics[width=1in,height=1.25in,clip,keepaspectratio]{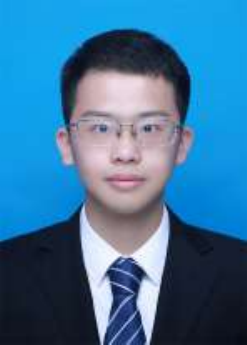}}]{Jintai Chen}
received the B.S. degree in applied statistics from the Zhongnan University of Economics and Law. He is currently working toward the Ph.D. degree in the College of Computer Science, Zhejiang University. His research interests includes deep learning, machine learning and data mining, especially on the computer vision, data mining, and medical intelligence. He has authored some papers at prestigious international conferences and journals, such as, ICML, AAAI, IJCAI, CVPR, MICCAI, and IEEE TMI, and served as reviewers for MICCAI, AAAI, IJCAI, IEEE TNNLS, and JBS.
\end{IEEEbiography}

\begin{IEEEbiography}[{\includegraphics[width=1in,height=1.25in,clip,keepaspectratio]{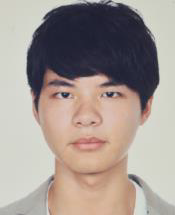}}]{Xiaojun Chen}
received the M.S. degree in applied informatics from the Liverpool University, UK, in 2017. His research interests includes machine learning, data mining, deep learning and computer vision. He has authored some papers at conferences and journals, such as, JACC and IEEE Access.
\end{IEEEbiography}

\begin{IEEEbiography}[{\includegraphics[width=1in,height=1.25in,clip,keepaspectratio]{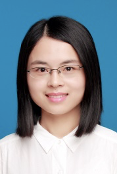}}]{Xiaoxian Yang}
received the Ph.D. degree in Management Science and Engineering from Shanghai University, China, in 2017. She is currently an assistant professor at Shanghai Polytechnic University, China. Her research interests include business process management, formal verification, wireless network, and mobile health. Her researches are supported by the National Natural Science Foundation of China (NSFC), Natural Science Foundation of Shanghai (NSFS), CERNET Innovation Project, and Foundation of Science and Technology Commission of Shanghai Municipality. She has published more than 20 papers in academic journals such as TITS, TOMM, TCSS, TOIT, FGCS, MONET, IJSEKE, FGCS, IJDSN, Remote Sensing, and international conferences such as CollaborateCom, SEKE and IDEAL. She obtained 2 patents and 3 registered software copyrights in China, involving Wireless Network, Workflow Management, and Formal Verification. Dr. Yang had participated in organizing international conferences and workshops, such as CollaborateCom 2018, Chinacom 2019 and Mobicase 2019. She also worked as Guest Editor for MONET and WINE, and served as reviewers for IEEE TII, IEEE T-ITS, ACM TOMM, ACM TOIT, Wiley ETT, FGCS, PPNA, JBHI, Wireless Networks, COMPUTER NETWORK, etc.
\end{IEEEbiography}

\begin{IEEEbiography}[{\includegraphics[width=1in,height=1.25in,clip,keepaspectratio]{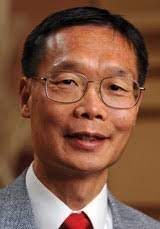}}]{Danny Z. Chen}
received the B.S. degrees in Computer Science and in Mathematics from the University of San Francisco, California, USA in 1985, and the M.S. and Ph.D. degrees in Computer Science from Purdue University, West Lafayette, Indiana, USA in 1988 and 1992, respectively. He has been on the faculty of the Department of Computer Science and Engineering, the University of Notre Dame, Indiana, USA since 1992, and is currently a Professor with tenure.  Dr. Chen's main research interests include computational biomedicine, biomedical imaging, computational geometry, algorithms and data structures, machine learning, data mining, and VLSI.  He has worked extensively with biomedical researchers and practitioners, published over 150 journal papers and 240 peer-reviewed conference papers in these areas, and holds 7 US patents for technology development in computer science and engineering and biomedical applications. He received the CAREER Award of the US National Science Foundation (NSF) in 1996, a Laureate Award in the 2011 Computerworld Honors Program for developing “Arc-Modulated Radiation Therapy” (a new radiation cancer treatment approach), and the 2017 PNAS Cozzarelli Prize of the US National Academy of Sciences. He is a Fellow of IEEE and a Distinguished Scientist of ACM.
\end{IEEEbiography}

\begin{IEEEbiography}[{\includegraphics[width=1in,height=1.25in,clip,keepaspectratio]{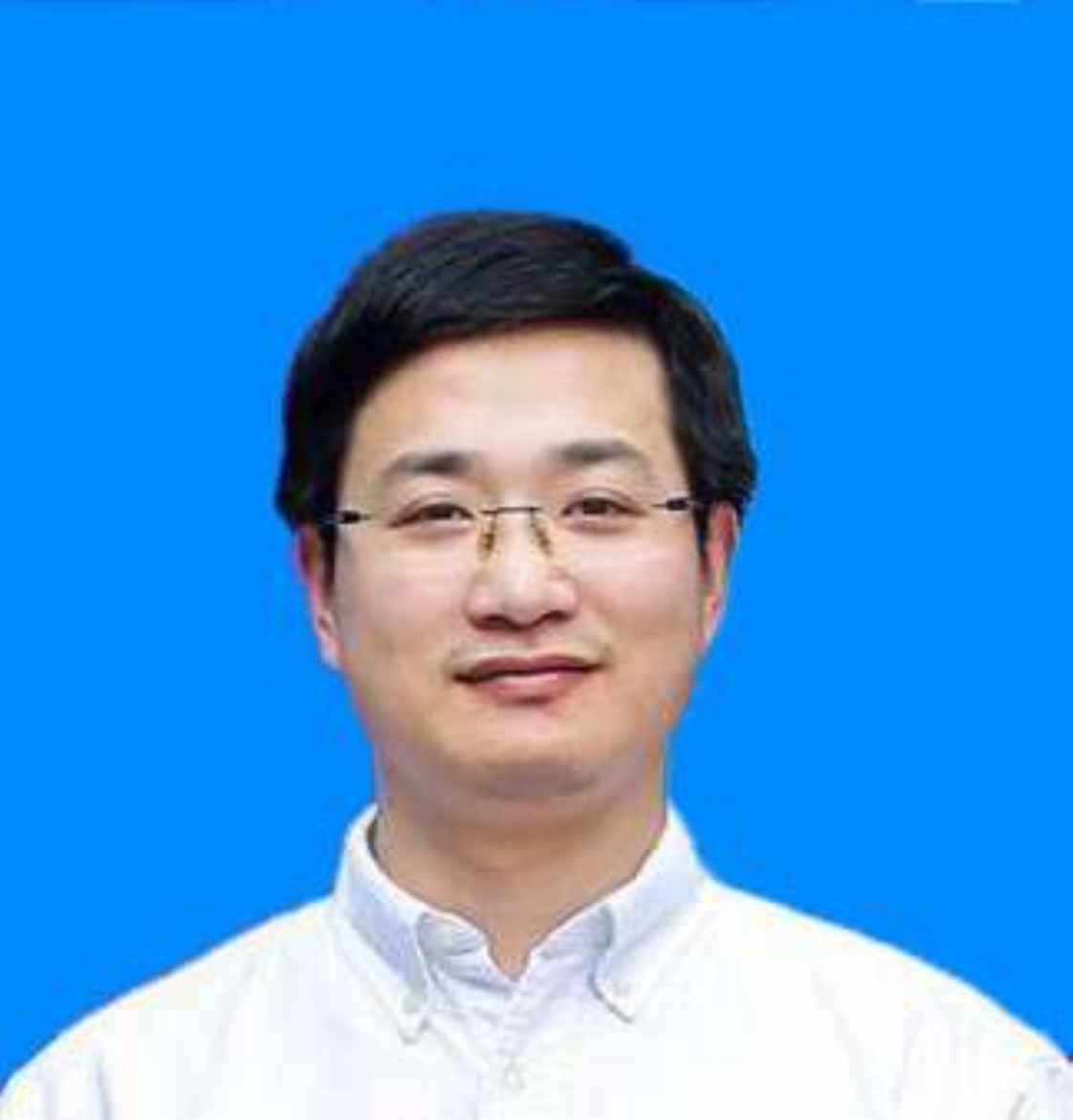}}]{Jian~Wu} received the Ph.D. degree in Computer Science and Technology from Zhejiang University in 1998. He is an IEEE member, CFF member, CCF TCSC member, CCF TCAPP member and member of the ``151 Talent Project of Zhejiang Province''. Prof. Wu is recently the director of  Research Centre of Zhejiang University and Vice-president of National Research Institute of Big Data of Health and Medical Sciences of Zhejiang University. His research interests include Medical Artificial Intelligence, Service Computing and Data Mining.
\end{IEEEbiography}

\clearpage

\end{document}